\documentclass{article}

\usepackage[final]{neurips_2023}

\usepackage{natbib}
\usepackage[utf8]{inputenc} % allow utf-8 input
\usepackage[T1]{fontenc}    % use 8-bit T1 fonts
\usepackage{hyperref}       % hyperlinks
\usepackage{url}            % simple URL typesetting
\usepackage{booktabs}       % professional-quality tables
\usepackage{amsfonts}       % blackboard math symbols
\usepackage{nicefrac}       % compact symbols for 1/2, etc.
\usepackage{microtype}      % microtypography
\usepackage{xcolor}         % colors
\usepackage{graphicx}
\usepackage{amsmath}

\title{Optimizing Likelihood-free Inference using Self-supervised Neural Symmetry Embeddings}

\author{%
  Deep~Chatterjee{$^{\dagger}$}\\
  Massachusetts Institute of Technology\\
  Cambridge, MA, USA
  \And
  Philip~C.~Harris{$^{*}$}\\
  Institute for Artificial Intelligence and Fundamental Interactions~(IAIFI)\\
  Massachusetts Institute of Technology\\
  Cambridge, MA, USA
  \And
  Maanas~Goel\\
  University of Pennsylvania \\
  Philadelphia, PA, USA\\
  \And
  Malina~Desai \\
  Massachusetts Institute of Technology \\
  Cambridge, MA, USA\\
  \And
  Michael~W.~Coughlin\\
  University of Minnesota \\
  Minneapolis, MN, USA \\
  \And
  Erik~Katsavounidis\\
  Massachusetts Institute of Technology\\
  Cambridge, MA, USA\\
}

\newcommand{\boldtheta}{\ensuremath{\mathbf{\Theta}}}
\newcommand{\data}{\ensuremath{\mathbf{d}}}
\newcommand{\bgamma}{\ensuremath{\mathbf{\Gamma}}}
\newcommand{\smallgamma}{\ensuremath{\mathbf{\gamma}}}
\newcommand{\vicreg}{\ensuremath{\mathcal{L}_{\mathrm{VICReg}}}}
\begin{document}

\maketitle

\begin{abstract}
  Likelihood-free inference is quickly emerging as a powerful tool to perform fast/effective parameter estimation.
  We demonstrate a technique of optimizing likelihood-free inference to make it even faster by marginalizing symmetries
  in a physical problem. In this approach, physical symmetries, for example, time-translation are learned using joint-embedding
  via self-supervised learning with symmetry data augmentations. Subsequently, parameter inference is performed using a normalizing flow where
  the embedding network is used to summarize the data before conditioning the parameters. We present this approach on two simple
  physical problems and we show faster convergence in a smaller number of parameters compared to a normalizing flow that does not use a pre-trained symmetry-informed
  representation.
\end{abstract}

\section{Introduction}
\label{sec:intro}
\renewcommand{\thefootnote}{$^{\dagger}$}
\footnotetext{\texttt{deep1018@mit.edu}; $^{*}$\texttt{pcharris@mit.edu}}
\renewcommand{\thefootnote}{\alph{footnote}}
\setcounter{footnote}{0}
Parameter estimation is fundamental to experimental science. Traditional
approaches involve stochastic sampling, like markov-chain monte carlo (MCMC) or
nested sampling, to obtain posterior probability densities and confidence intervals on parameters.
However, with the increase in number and flavors of analyses, complexity of experiments, and the demand for
near real-time inference, stochastic sampling is turning out to be a challenge both in terms of time and required compute.
Likelihood-free, or simulation-based inference (LFI/SBI) has emerged as a potential breakthrough in this regard
(see~\cite{Cranmer_2020,papamakarios2021normalizing} for a review). Simulations to generate data,
\data, corresponding to physical parameters, \boldtheta, exists for many experiments. The idea of
LFI is to learn $p(\boldtheta, \data)$, $p(\boldtheta \vert \data)$, or $p(\data \vert \boldtheta)$ from the
pairs $\{\boldtheta_i, \data_i\}$ as desired (see Sec. 2.1 of~\cite{Alsing_2019} for a nice description). In
this work, we focus on the construction of the posterior, $p(\boldtheta\vert\data)$.

A critical aspect of the application of LFI arises from the difficulty in performing high dimensional fits. Often times, posterior extraction requires a large amount of training samples, and large neural network to ensure a fully accurate description of the model. As an example, current state of the art networks for gravitational-wave (GW) parameter estimation require over 100 million parameters to ensure an accurate description of the GW parameters~\citep{Dax_2021}.
One potential way to mitigate the size of LFI models is to project out parameters that are known to be invariant to the desired intrinsic parameters.
For example, in the case of pulse-like time-series, the time of arrival may not be as important as the physical system/process that produces the pulse. 
However, a conventional application of LFI, unaware of this symmetry, considers data corresponding to time translated signals as \emph{unique} and requires that the arrival time be determined, or marginalized over, with the other desired physical parameters.
This necessitates a larger
model and expensive compute to train a model that learns to marginalize over time (or some other variables, $\pmb{\lambda}$), despite the fact that the desired parameter inference is time-(or $\pmb{\lambda}$-) invariant. If the interest is limited to the physical, intrinsic parameters, this is superfluous.

In this paper, we utilize self-supervision to project away the invariant parameters so as to reduce the LFI to the extraction of only the intrinsic parameters. Here, we marginalize over this symmetry by learning a \emph{similarity} embedding through self-supervised learning (SSL). In the instance of time translation invariance we rely on SSL to build a reduced representation where instances of data with different arrival times are represented as similar values in an embedded space. We train a representation network by minimizing the VICReg similarity loss~\citep{bardes2022vicreg}. A similar technique was used recently to summarize cosmological data \citep{akhmetzhanova2023data}.
SSL has also been applied to learn symmetries in partial different equations (PDEs)~\citep{mialon2023selfsupervised}. There are other techniques, not involving SSL to summarize data for LFI on timeseries e.g. YuleNet~\citep{rodrigues2020learning}. 
Once we have constructed the SSL space, the subsequent LFI is conditioned on the embedded space. We demonstrate this scheme considering two simple physical systems -- a damped harmonic oscillator (subsequently called SHO)
that is started at different times, and a sine-gaussian pulse (subsequently called SG) that has an arbitrary time of arrival in a window. We show that
we are able to obtain posteriors which are accurate and require a significantly smaller number of training parameters compared
to the case when the data is not represented, or similarity pre-training is not performed.
We point out that while \emph{signals} with identical physical parameters, translated in time is an exact symmetry, the corresponding \emph{data} is
not due to the presence of noise. This motivates ``similar'' representation as opposed to ``same.''

\section{Marginalizing Time-translation in Physical problems}
\label{sec:time_translation}
\begin{figure}
    \centering
    \includegraphics[width=0.4\textwidth, trim=0cm 0.4cm 1cm 0cm, clip]{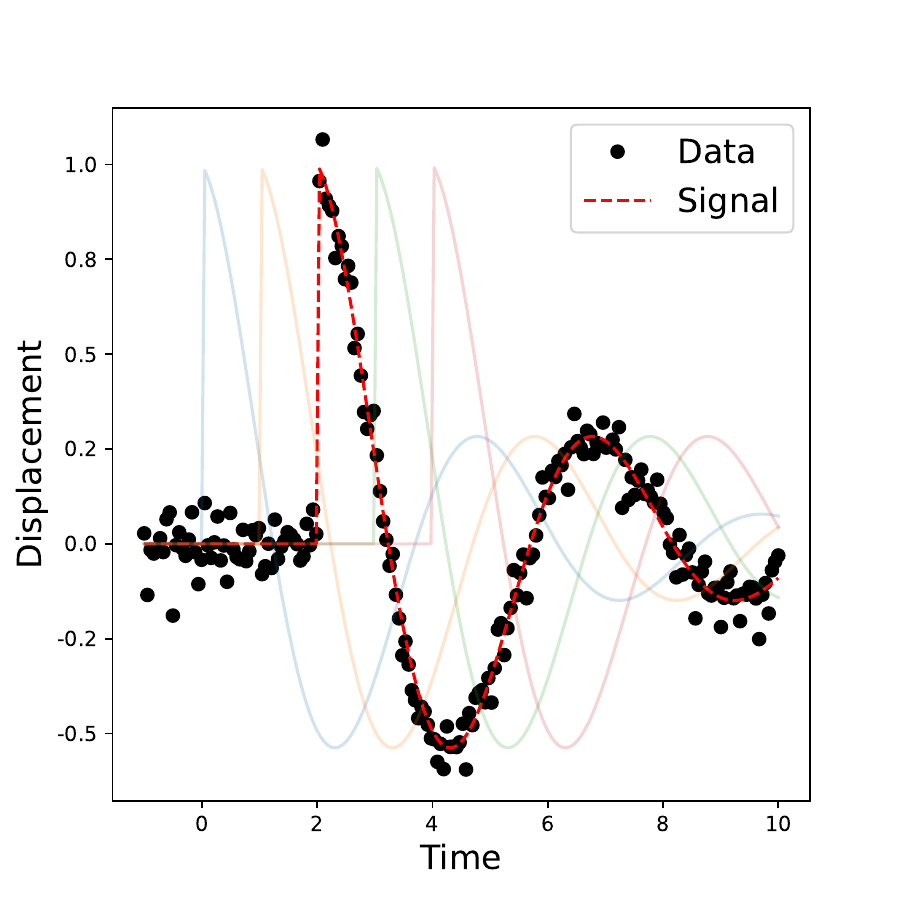}\hspace{0.5cm}
    \includegraphics[width=0.4\textwidth, trim=0cm 0.4cm 1cm 0cm, clip]{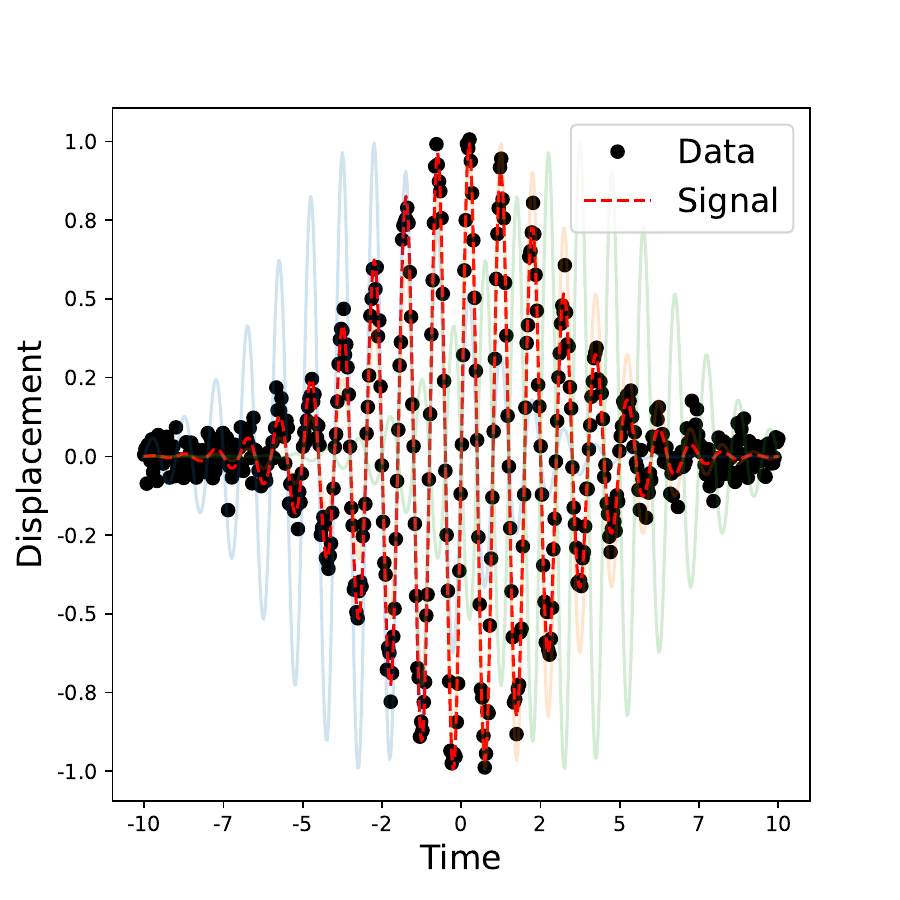}
    \caption{\textbf{Left:} Example signal and data of a damped harmonic oscillator (SHO) with
    representative values of $\boldtheta = \{\omega_0, \beta\}$. The data
    is shown by the solid dots to represent sampling a time-series.
    \textbf{Right}: An example of a sine-gaussian (SG) pulse with parameters, $\boldtheta=\{f_0, \tau\}$.
    In either case, different times of arrival, shown in faded curves, do not affect the inference of
    parameters.}
    \label{fig:sho_data_signal}
\end{figure}
Consider a SHO  described by physical parameters, $\boldtheta = \{\omega_0, \beta\}$,
the natural frequency and damping coefficient respectively. We would like to infer on the parameters given a time-series, \data, with SHO amplitudes as shown in the left panel of Fig.~\ref{fig:sho_data_signal}; white noise is additionally added. If the oscillator is
started at a different time, represented by the shifted solid lines in the figure, the inference on $\omega_0$ and $\beta$ remains the
same. The same applies in the context of inferring properties of a SG pulse, represented in the right panel of Fig.~\ref{fig:sho_data_signal},
where the parameters of interest are the central frequency, $f_0$, and width,
$\tau$. In the standard scheme of LFI, the parameters are sampled from prior distribution, $\boldtheta \sim p(\boldtheta)$, following which
the data, $\data$, is generated, and the model is provided $\{\boldtheta_i, \data_i\}$ pairs. This, however, leaves learning nuisance
parameters, like time shifts, and marginalizing over them, up to the model, which could be expensive in terms of trainable parameters
and compute. 

\subsection{Representing time-shifted data}
\label{subsec:representation}
\begin{figure}
    \centering
    \includegraphics[width=0.4\textwidth, trim=0cm 0cm 0.3cm 0cm, clip]{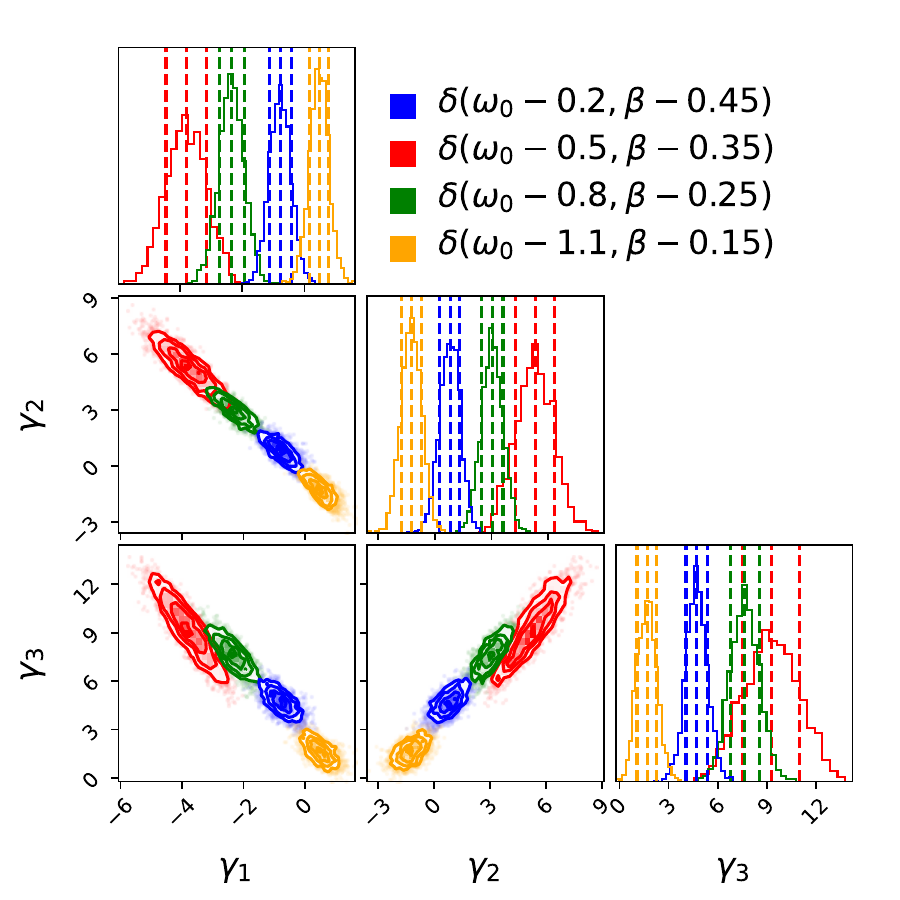}
    \includegraphics[width=0.4\textwidth, trim=0cm 0cm 0.3cm 0cm, clip]{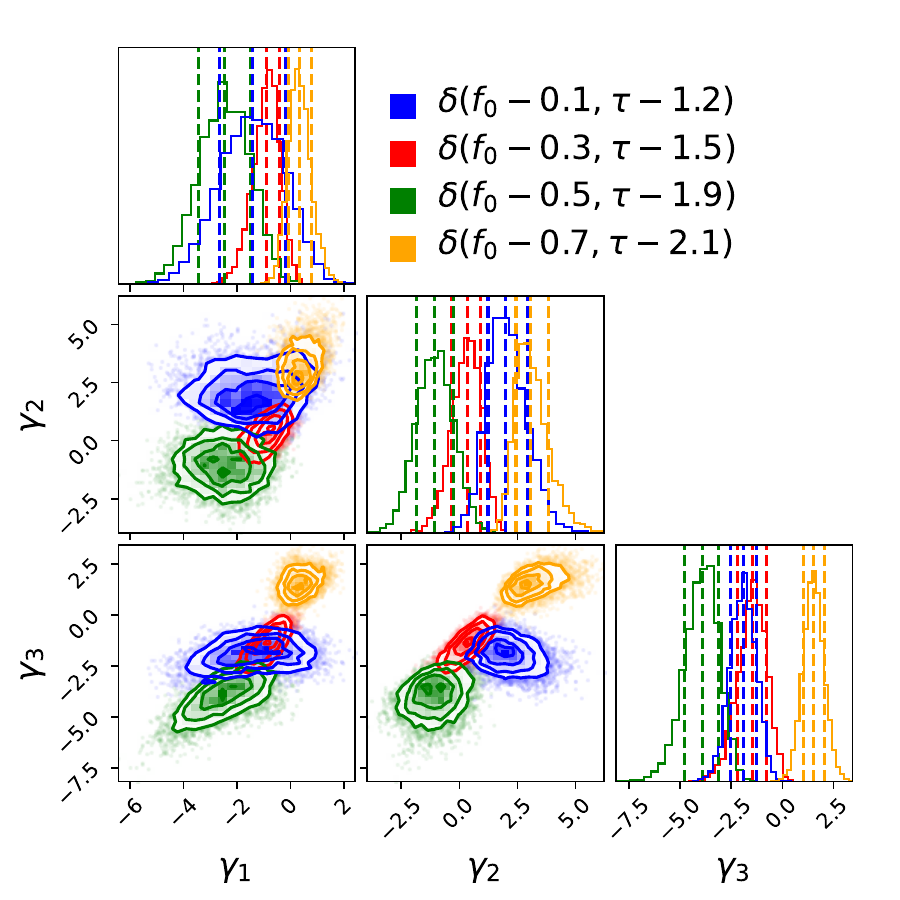}
    \caption{SHO (left) and SG (right) data representations. The signals have intrinsic
    parameters denoted by the delta function, but have augmented time shifts
    sampled uniformly from a window. We observe that the SSL construction leads to separation based on intrinsic parameters where the variations in the time of arrival is reduced to a substantially smaller region that the intrinsic parameters, approaching the time-project space.}
    \label{fig:reps}
\end{figure}
We project the time shifted data instances into a 3D embedded space via a neural network, $\bgamma \equiv h\circ f$, such that
the data at different times, with the same parameters $\boldtheta$, is mapped to similar regions in this embedded space. 
We follow the scheme of self-supervised learning as follows:
\begin{itemize}
    \item We consider two batches -- in the first case the data, \data, has a fixed reference time of arrival, in the second,
          the data, $\data'$, is augmented by shifting start time by a suitable time-shift prior.
    \item In our implementation, $f$ is a 1D convolutional resnet followed by a fully connected contraction layer to represent each
          batch in 3 dimensions i.e. $\smallgamma\in\mathbb{R}^3$. The representation is then passed through a fully-connected expander network, $h$, to output $x\in\mathbb{R}^{12}$ i.e.,
          \begin{equation}
            \smallgamma = f(\data);\;\smallgamma' = f(\data');\;x = h(\smallgamma);\;x' = h(\smallgamma').
            \label{eq:representation}
          \end{equation}
          Ideally, the dimensionality of the representation is determined based on a systematic hyper-parameter tuning. However, in this case, our choices dimensions was based on performance from trials on a handful of cases.
    \item We use the VICReg loss between the expanded outputs to minimize regularized variance and covariance of the embedded batches individually, and the invariance between the batches using mean-squared error. The net loss is a weighted sum of the three terms,
          \begin{eqnarray}
              \vicreg(x, x') = \lambda_1\;\mathrm{MSE}(x, x') + 
              \lambda_2\;\left[\sqrt{\mathrm{Var}(x) + \epsilon} + \sqrt{\mathrm{Var}(x') + \epsilon}\right] + \nonumber \\
              \hspace{5cm}\lambda_3\;\left[C(x) + C(x')\right].
          \label{eq:loss}
          \end{eqnarray}
          Here $C$ is the squared sum of off-diagonal elements of the covariance matrix, scaled by the
          number of dimensions (see Eq.~4 of \cite{bardes2022vicreg}), and $\lambda_{1,2,3}$ are the relative weighting of the three terms
          that can be adjusted during training. For this work we found that increasing the reconstruction and variance weights compared to
          the covariance term in the initial epochs, followed by setting the weights equal helped with the convergence.
    \item Once trained, data from same physical parameters are represented as clusters embedded in this space irrespective
          of the time of arrival. This is shown in Fig.~\ref{fig:reps}, where different physical parameters are represented
          as roughly separated clusters irrespective of their time of arrival. We freeze part of the network $f$ at this
          point.\footnote{Here, we freeze the convolutional layers of $f$, and left the fully connected layers free
          to train in the next step.}
\end{itemize}

\section{LFI with Normalizing Flows}
\label{sec:lfi}
\begin{figure}
    \centering
    \includegraphics[width=0.48\textwidth, trim=0.4cm 0.5cm 0cm 0.2cm, clip]{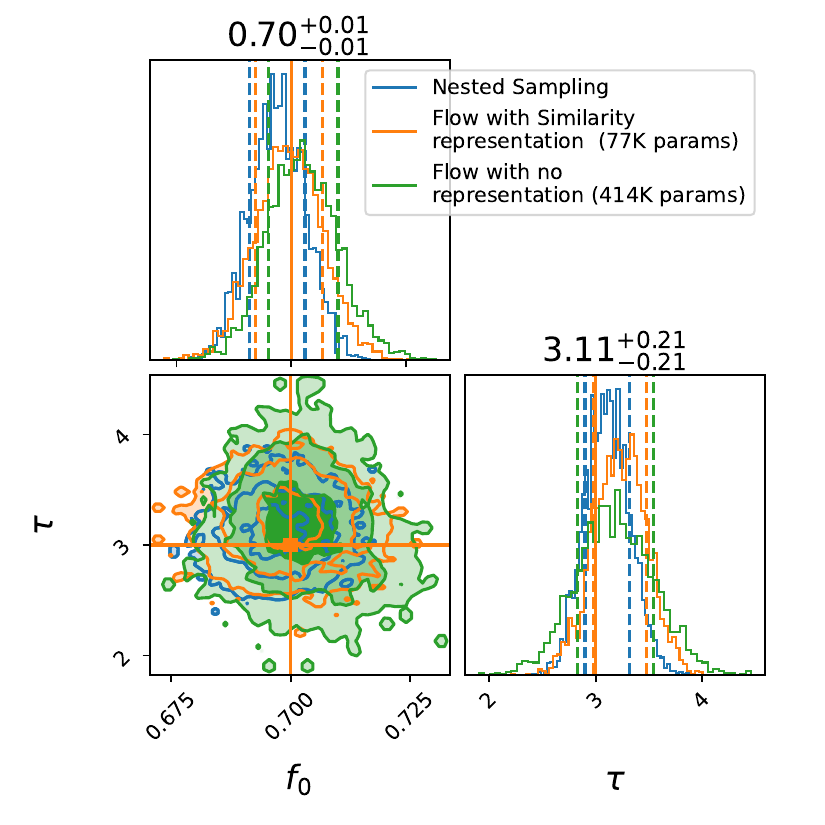}
    \includegraphics[width=0.48\textwidth, trim=0.4cm 0.5cm 0cm 0.2cm, clip]{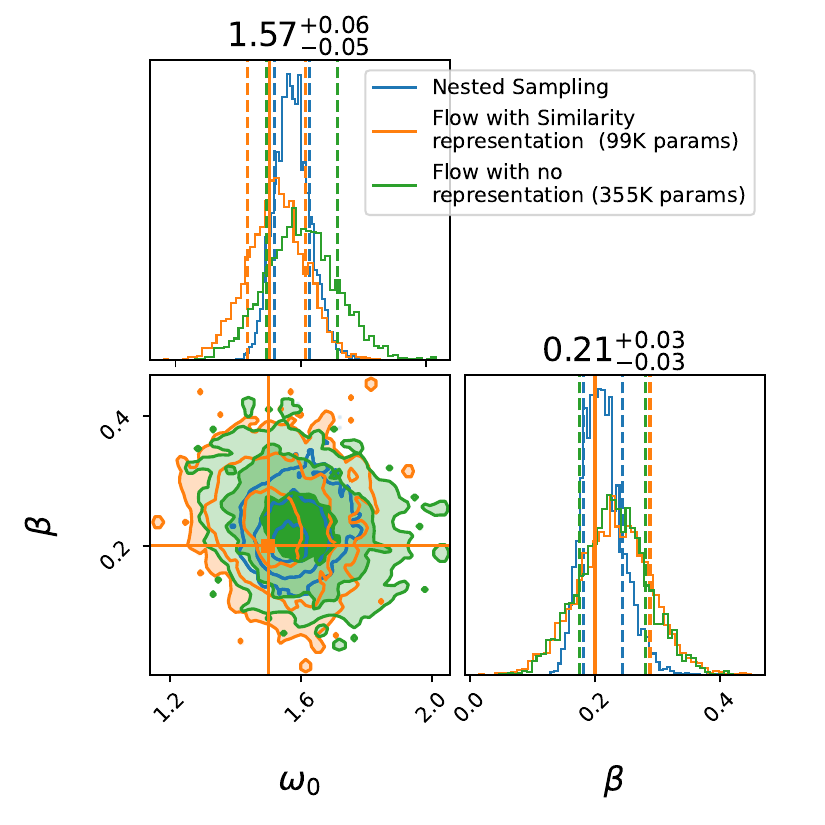}
    \\
    \includegraphics[width=0.48\textwidth, trim=-0.5cm 0.5cm 0cm 0.9cm,clip]{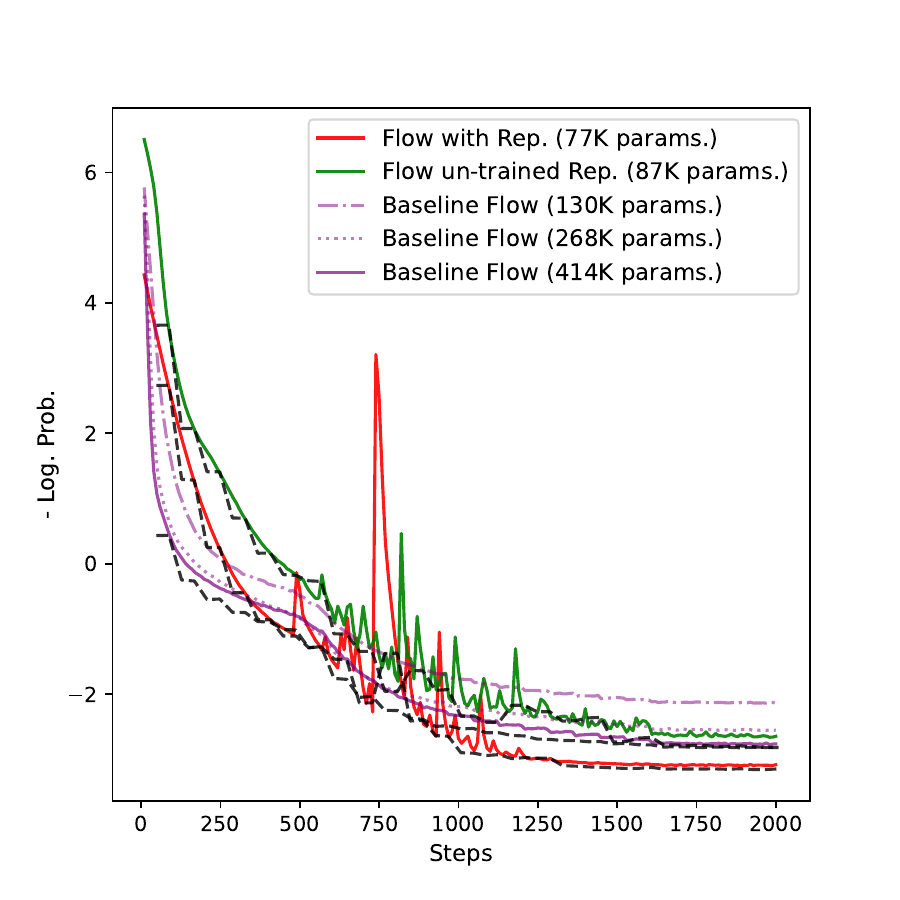}
    \includegraphics[width=0.48\textwidth, trim=-0.5cm 0.5cm 0cm 0.9cm,clip]{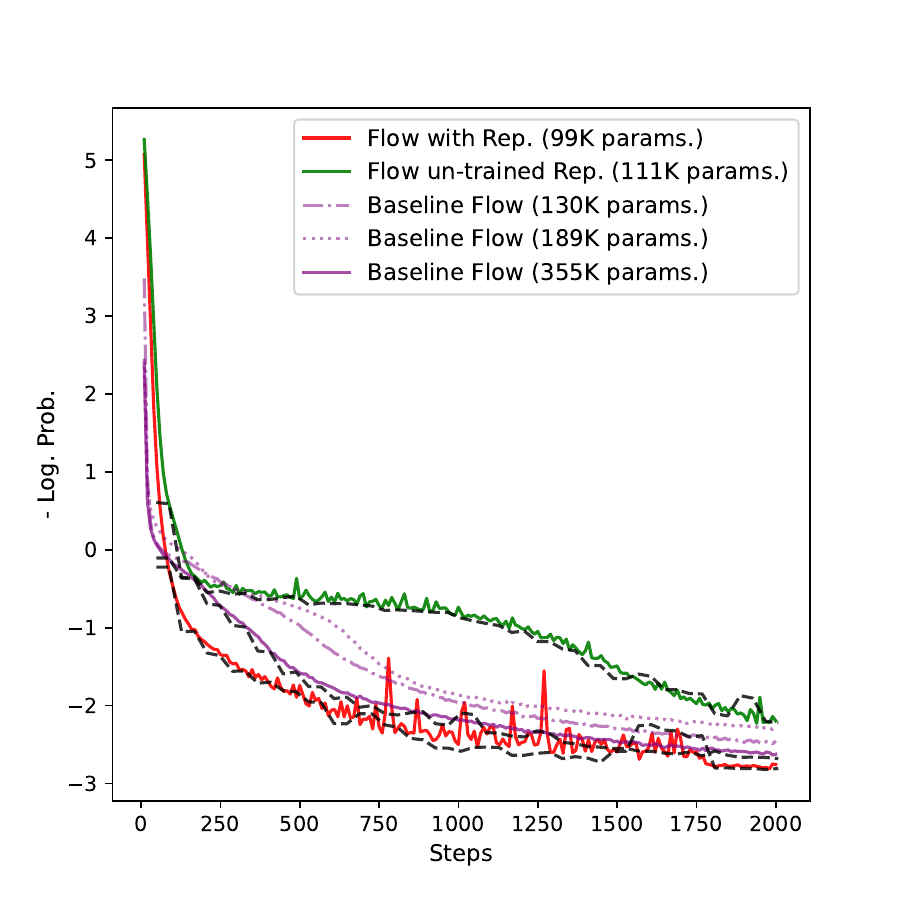}
    \caption{\textbf{Top}: We show consistency between posteriors from a NF with a pre-trained \bgamma, a baseline NF
    without any data summary, and results from nested sampling. The $\pm$ uncertainties in the individual panels are quoted
    based on the nested sampling result. \textbf{Bottom}: Loss curves (black dashed lines
    for validation) for the different cases. The NF with a pre-trained {\bgamma} converges in a smaller number of trainable
    parameters, and sometimes more rapidly, compared to other cases.}
    \label{fig:posterior_plot}
\end{figure}
\begin{figure}
    \centering
    \includegraphics[width=0.48\textwidth, trim=-0.5cm 0.5cm 0cm 0.2cm,clip]{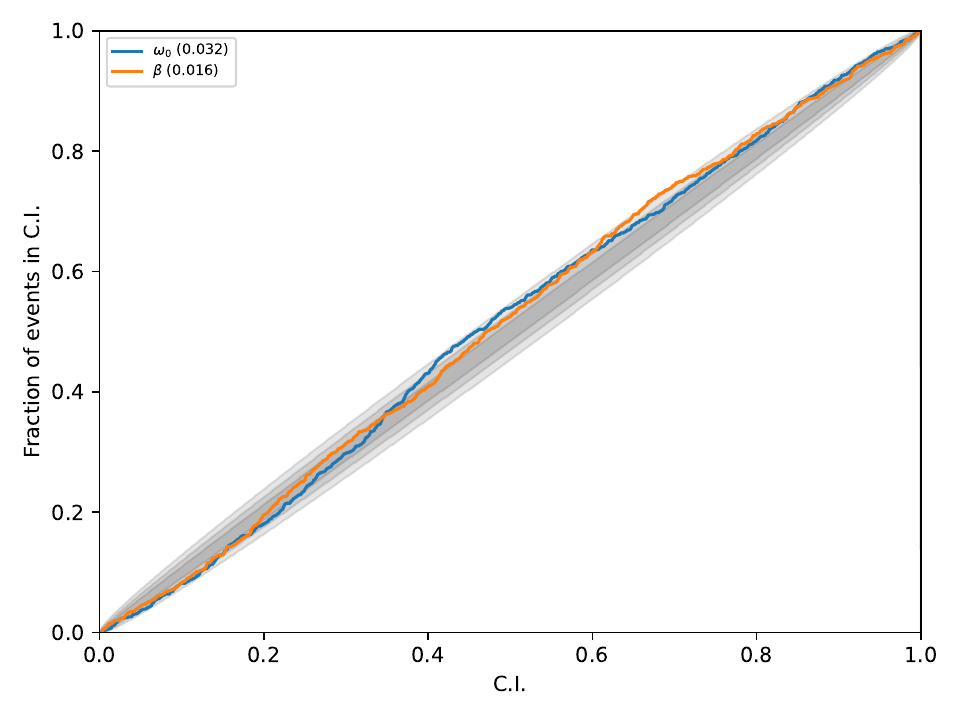}
    \includegraphics[width=0.48\textwidth, trim=-0.5cm 0.5cm 0cm 0.2cm,clip]{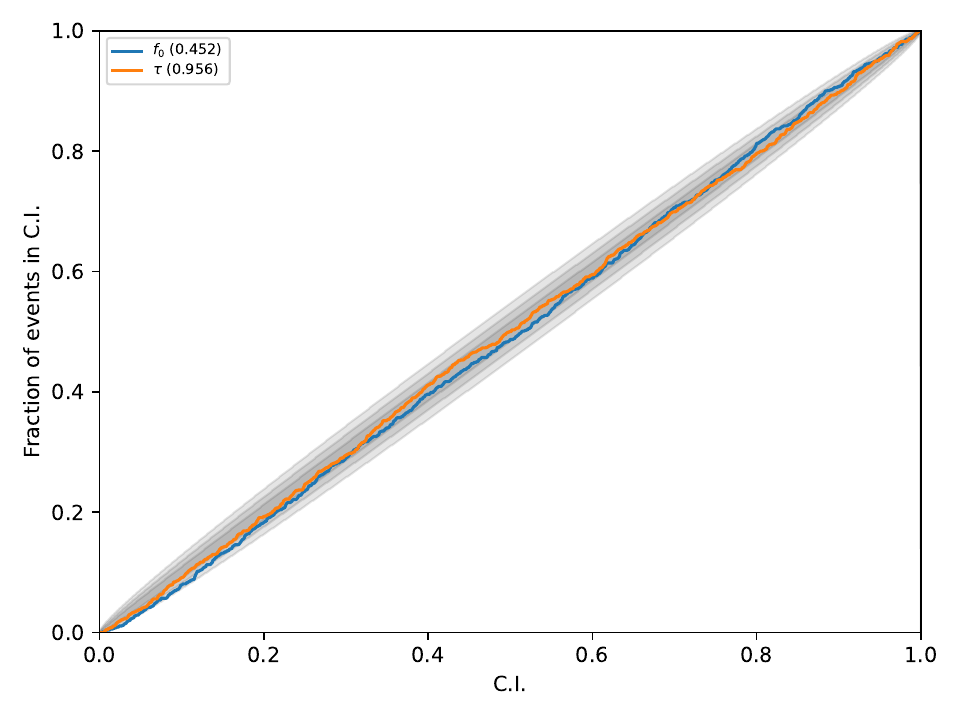}
    \caption{\textbf{Left}: Results from 1000 tests data instances for which posteriors where drawn. It plots
    fraction of times the true parameters of the SHO model lie in credible interval. In each case 3000
    posterior samples are drawn from the trained model. The gray bands denote the 1-, 2-, 3-$\sigma$ confidence intervals in
    decreasing order of opacity.
    \textbf{Right}: Same plot as the left for the SG model and testing set.}
    \label{fig:pp_plots}
\end{figure}
Next, we perform the conventional LFI by training a normalizing flow (NF)~\citep{rezende2016variational}. We use a base 2D standard normal
distribution with affine autoregressive transforms i.e. a masked autoregressive flow (MAF)~\cite{papamakarios2018masked}.\footnote{
We use the implementation from {\tt{nflows}} library with minor changes.}
We condition the normalizing flow with the representation network, $f$. Hence, the pairs provided during training are
$\{\boldtheta_i, f(\data_i)\} \equiv \{\boldtheta_i, \pmb{\gamma}_i\}$. The network is trained by maximizing the
log-likelihood. In Fig.~\ref{fig:posterior_plot} we show an example of posterior samples from testing data for both SHO and
SG models. We compare the same with posteriors samples obtained by training a baseline NF which does not have any such
representation network, and also with results from nested sampling.\footnote{We use the {\tt{bilby}} inference library with
{\tt{dynesty}} sampler.} We compare the widths of the posterior obtained using our trained model and stochastic sampling and show that they are consistent. We also compute the Cramer-Rao bound on the parameter uncertainty by expanding the likelihood to quadratic order in the argument of the exponential. We show that the results obtained from the trained flow are consistent with CRB for the SG and SHO models (see Appendix~\ref{appendix:analytic}). 
In Fig.~\ref{fig:pp_plots} we show the result of 1000 test data instances for which posteriors were drawn from the trained model.
The plot highlights the consistency between the confidence intervals and the fraction of times the true parameters lie inside the confidence intervals. The diagonal trend in the figure, along with the consistency of individual posteriors mentioned in Appendix~\ref{appendix:analytic} show consistency across the entire testing dataset. The gray bands on both panels denote the 1-, 2-, 3-$\sigma$ confidence intervals.

Comparing number of trainable parameters -- the baseline NF model, without any data summary, has $\mathcal{O}(10^6)$ while
the same for the NF with pre-trained {\bgamma} is significantly lower i.e. $\mathcal{O}(10^5)$. For a batch
size of $10^3$, a forward pass involves $\sim 1.1\times10^8$ multiply–accumulate operations (MACs)
for the NF with the representation, while $\sim 4.1\times10^8$ MACs for the baseline NF. This may translate to longer
training times, for example, timing 100 epochs a NVIDIA GeForce GTX 1080 Ti GPU with SGD optimizer takes $\sim 20$ minutes
for the NF with representation while $\sim 80$ minutes for the baseline NF. We found 100 epochs of the similarity pre-training,
which takes $\sim 10$ minutes, was sufficient to reduce the individual loss terms in Eq.~(\ref{eq:loss}). We found that
freezing all trainable parameters of representation network parameters at this stage led to slightly larger posterior widths compared to nested sampling.
However, letting the network further train with the flow, at least leaving the fully connected layers of the representation network unfrozen, led to
similar performance as nested sampling.
Therefore, the prescription presented here
generally implies smaller networks and more efficient memory and compute requirements.

\section{Application in Physics and Astronomy}
While in this work, we have considered projecting out the time of arrival as a specific example of a symmetry that does not impact the measurement of other parameters, it should be noted that the technique presented here could be used in general to marginalize over parameters that are not of interest. Many problems physics and astronomy exhibit a set of \emph{intrinsic} vs. \emph{extrinsic} parameters. For example, in GW
data analysis, the intrinsic parameters include the masses and spins of merging binary black holes, while extrinsic parameters
involve distance, inclination, sky-location etc. Likewise, a kilonova lightcurve is intrinsically characterized by the ejecta mass and lanthanide fraction etc., and extrinsically by redshift, extinction, time of arrival etc. If the interest is only limited to a subset of parameters, or to only intrinsic properties, a representation using self-supervised learning can be used for efficient LFI.

This work is supported by NSF HDR Institute Grant PHYS-2117997, “Accelerated AI Algorithms for Data-Driven Discovery.”

\appendix
\newpage
\section{Appendix}
\subsection{Analytic approximation of the posterior}\label{appendix:analytic}
The signal models are,
\begin{align}
  y_i \equiv y(t_i) = 
  \begin{cases}
    \exp\left(-\beta \omega_0 t_i\right) \;\cos \left(\omega_0 t_i\sqrt{1-\beta^2} \right) & \text{for SHO}\\
    \exp\left(-\frac{t_i^2}{\tau^2}\right) \;\sin (2\pi f_0 t_i), & \text{for SG}.
  \end{cases}
  \label{eq:signal_models}
\end{align}
Assuming white noise drawn from $\mathcal{N}(0, \sigma)$, (with $\sigma$ known) the likelihood is,
\begin{align}
    \mathcal{L} \propto \exp\left[-\frac{1}{2}\sum_i \left(\frac{y_i - \hat{y}_i}{\sigma}\right)^2\right],
\end{align}
where $\hat{y}$ is the truth i.e. signal corresponding to true parameters $\{\hat{\omega}_0,\hat{\beta}\}$
($\{\hat{f}_0,\hat{\tau}\}$) for SHO (SG).
Expanding the argument of the exponential to quadratic order in $\{{\omega_0}, \beta\}$ (or $\{{f}_0,{\tau}\}$)
around the true value, the likelihood is approximated by gaussian with widths,
\begin{subequations}
\begin{align}
\Delta\omega_0 &\approx \frac{\sigma}{\sqrt{\sum_i^N t_i^2 e^{-2 \hat{\beta}\hat{\omega}_0 t_i}\left[ \sqrt{1-\hat{\beta}^2} \sin \left( \hat{\omega}_0 t_i\sqrt{1-\hat{\beta}^2} \right) + \hat{\beta}\cos \left(\hat{\omega}_0 t_i\sqrt{1-\hat{\beta}^2}\right)\right]^2}},
\\
\Delta\beta &\approx \frac{\sigma}{\sqrt{\sum_i^N \hat{\omega}_0^2 t_i^2 e^{-2 \hat{\beta}\hat{\omega}_0 t_i}\left[ \frac{\hat{\beta}\sin \left( \hat{\omega}_0 t_i\sqrt{1-\hat{\beta}^2} \right)}{\sqrt{1-\hat{\beta}^2}} - \cos \left(\hat{\omega}_0 t_i\sqrt{1-\hat{\beta}^2} \right) \right]^2}},
\\
\Delta\tau &\approx \frac{\sigma \hat{\tau}^3}{\sqrt{\sum_i^N 4\;t_i^4\exp(-2t^2_i/\hat{\tau}^2)\sin^2(2\pi \hat{f}_0 t_i)}},
\\
\Delta f &\approx \frac{\sigma}{\sqrt{\sum_i^N 4\pi^2 \;t_i^2\exp(-2t^2_i/\hat{\tau}^2)\cos^2(2\pi \hat{f}_0 t_i)}}.
\end{align}    
\label{eq:crb}
\end{subequations}
By assuming flat priors, the posterior is proportional to the likelihood, having widths approximately given
by Eqs.~(\ref{eq:crb}). For the examples in Fig.~\ref{fig:posterior_plot}, the value of $\sigma = 0.4$ in either case.
The true SG parameters are $\{\hat{f}_0 = 0.7, \hat{\tau}=0.3\}$; for SHO they are $\{\hat{\omega}_0 = 1.5, \hat{\beta}=0.2\}$.
From Eqs.~(\ref{eq:crb}), we get $\Delta\ f_0 \approx 0.006\;\Delta\tau\approx 0.2$ for SG,and $\Delta\omega_0 \approx 0.04\;\Delta\beta\approx 0.03$
for SHO models, satisfying Cramer-Rao bound for the one-sigma widths shown in Fig.~\ref{fig:posterior_plot}.

\subsection{Code Availability}
The codes used in the analysis, including trained weights to reproduce figures in the draft, are provided publicly and \href{https://github.com/ML4GW/summer-projects-2023/releases/tag/neurips-2023}{tagged}. Online rendering of notebooks are available here: \url{https://github.com/ML4GW/summer-projects-2023/tree/neurips-2023/symmetry-informed-flows}.
%%%%%%%%%%%%%%%%%%%%%%%%%%%%%%%%%%%%%%%%%%%%%%%%%%%%%%%%%%%%
\bibliographystyle{abbrvnat}
\bibliography{references}
\end{document}